\documentclass[conference]{IEEEtran}
\usepackage{amsmath}
\usepackage{amsfonts}
\usepackage{multirow}
\usepackage{multicol}
\usepackage{graphicx}
\usepackage{footnote}
\usepackage{times}
\usepackage{textcomp}
\makesavenoteenv{tabular}
\makesavenoteenv{table}
\usepackage{algorithmicx}
\usepackage{algpseudocode}
\usepackage{algorithm}
\usepackage{cite}
\usepackage{pifont}
\usepackage{url}
\usepackage[table]{xcolor}

\definecolor{light-gray}{gray}{0.9}

\ifCLASSINFOpdf
\else
\fi
\begin{document}
%
\title{Identifying Irregular Power Usage by Turning Predictions into Holographic Spatial Visualizations}


%
\author{\IEEEauthorblockN{Patrick Glauner\IEEEauthorrefmark{1},
Niklas Dahringer\IEEEauthorrefmark{1},
Oleksandr Puhachov\IEEEauthorrefmark{1},
Jorge Augusto Meira\IEEEauthorrefmark{1}, \\
Petko Valtchev\IEEEauthorrefmark{2},
Radu State\IEEEauthorrefmark{1} and
Diogo Duarte\IEEEauthorrefmark{3}}
\IEEEauthorblockA{\IEEEauthorrefmark{1}Interdisciplinary Centre for Security, Reliability and Trust, University of Luxembourg\\
2721 Luxembourg, Luxembourg\\
Email: \{first.last, jorge.meira\}@uni.lu}
\IEEEauthorblockA{\IEEEauthorrefmark{2}Department of Computer Science, University of Quebec in Montreal, \\
H3C 3P8 Montreal, Canada\\
Email: valtchev.petko@uqam.ca}
\IEEEauthorblockA{\IEEEauthorrefmark{3}CHOICE Technologies Holding S\`arl\\
2453 Luxembourg, Luxembourg \\
Email: diogo.duarte@choiceholding.com}
}


\maketitle

\begin{abstract}
Power grids are critical infrastructure assets that face non-technical losses (NTL) such as electricity theft or faulty meters. NTL may range up to 40\% of the total electricity distributed in emerging countries. Industrial NTL detection systems are still largely based on expert knowledge when deciding whether to carry out costly on-site inspections of customers. Electricity providers are reluctant to move to large-scale deployments of automated systems that learn NTL profiles from data due to the latter's propensity to suggest a large number of unnecessary inspections.
In this paper, we propose a novel system that combines automated statistical decision making with expert knowledge. First, we propose a machine learning framework that classifies customers into NTL or non-NTL
using a variety of features derived from the customers' consumption data. The methodology used is specifically tailored to the level of noise in the data. Second, in order to allow human experts to feed their knowledge in the decision loop, we propose a method for visualizing prediction results at various granularity levels in a spatial hologram. Our approach allows domain experts to put the classification results into the context of the data and to incorporate their knowledge for making the final decisions of which customers to inspect.
This work has resulted in appreciable results on a real-world data set of 3.6M customers. Our system is being deployed in a commercial NTL detection software.
\end{abstract}


%
\IEEEpeerreviewmaketitle

\section{Introduction}
Critical infrastructure refers to assets that are essential for the functioning of a society and economy. They include power generation, transmission and distribution facilities. Losses in power grids can be grouped into technical losses, which appear naturally due to internal electrical resistance, and non-technical losses (NTL), which appear during power distribution. NTL include, but are not limited to, the following causes \cite{chauhan2013non, glauner2017challenge}:
\begin{itemize}
\item Meter tampering in order to record lower consumptions
\item Bypassing meters by rigging lines from the power source
\item Arranged false meter readings by bribing readers
\item Faulty or broken meters
\end{itemize}
NTL can range up to 40\% of the total electricity distributed in countries such as Brazil, India, Malaysia or Lebanon \cite{depuru2013high}. As a consequence, electricity providers face financial losses as well as a decrease of stability and reliability in their power networks. It is for these reasons that electricity providers aim to reduce NTL in their networks by carrying out on-site inspections of customers that have potentially irregular behavior. 
To date, most NTL detection systems deployed in industry are based on expert knowledge rules \cite{glauner2017challenge}. In contrast, the predominant research direction reported in the recent research literature is the use of machine learning/data mining methods, which learn from customer data and known irregular behavior that was reported through inspection results. Due to the high costs per inspection and the limited number of possible inspections, electricity providers aim to maximize the return on investment (ROI) of inspections.

\begin{figure}[!t]
\centering
\includegraphics[width=0.425\textwidth]{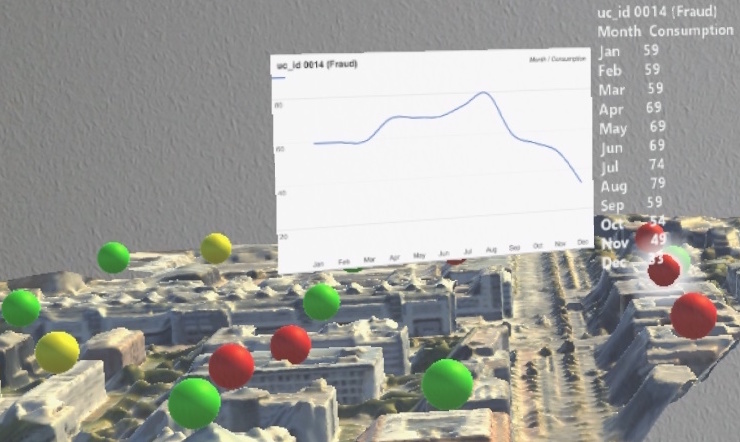}
\caption[XXX]{Example usage of our NTL detection system: Customers are classified as either regular (green), irregular (red) or suspicious (yellow) by a machine learning system. Holographic spatial visualization of customers allows domain experts at the electricity providers to gather information about the customers as well as their neighborhood in order to decide which customers to inspect. The figure depicts the profile of an irregular customer whose consumption has significantly dropped in the last few months.}
\label{fig:example}
\end{figure}

In this paper, we combine both worlds in a spatiotemporal approach that allows domain experts to visualize the prediction results of NTL classifiers in a holographic spatial visualization. An example of this outcome is depicted in Figure~\ref{fig:example}.

\begin{figure}[!t]
\centering
\includegraphics[width=0.5\textwidth]{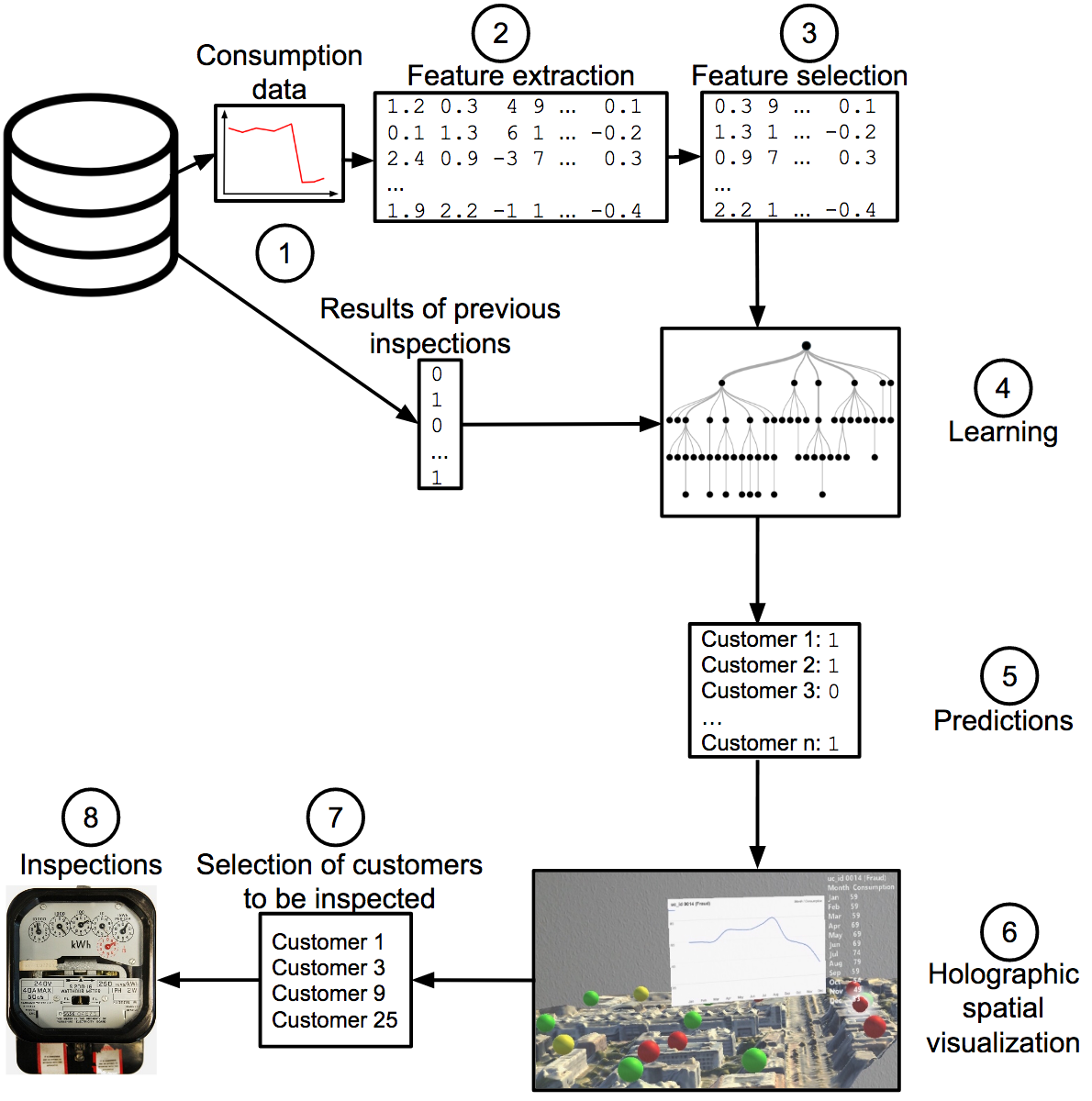}
\caption[XXX]{Proposed NTL detection system: First, the data of previously inspected customers is loaded, which consists of their consumption data as well the inspection result. Second, a vast number of features are extracted from the customers' noisy consumption data. Third, these features are reduced in order to retain the statistically meaningful ones. Fourth, using the set of reduced features and the results of previously carried out inspections, classifiers are trained in order to recognize NTL. Fifth, these classifiers are then used to predict for customers whether they should be inspected for NTL or not. Sixth, domain experts visualize the customers, their neighbors, inspection results and other data such as the consumption data in a spatial 3D hologram. Seventh, using their expert knowledge, they can review and amend the recommendations made by the classifiers in order to choose the customers for which an inspection appears to be justified from an economic point of view. Last, the inspections are carried out by technicians.}
\label{fig:process}
\end{figure}

\newpage
The main contributions of this paper are:
\begin{itemize}
\item We propose a novel and flexible framework to compute a large number of domain-specific features and generic features from the noisy consumption time series of customers for NTL detection.
\item We retain the statistically meaningful features extracted from the noisy consumption data and optimize different classifiers to predict NTL.
\item We present a novel approach to put the prediction results into context by visualizing them in a 3D hologram that contains information about customers and their spatial neighborhood. This hologram can be visualized in a Microsoft HoloLens.
\end{itemize}
The entire process in our proposed NTL detection system is depicted in Figure~\ref{fig:process}. As an outcome, domain experts can put the results generated by the classifiers into the context of the data in order to make the final decisions of whether to inspect specific customers.
We are confident that this approach will lead to an increase of both stability and reliability of power grids by making better use of the limited number of inspections as well as lead to a greater ROI of the limited number of inspections. 

\section{Related Work}
\label{chapter:review}
State-of-the-art surveys on NTL detection are provided in \cite{viegas2017solutions, glauner2017challenge}.
A data set of \texttildelow 22K customers is used in \cite{costa2013fraud} for training a neural network. It uses the average consumption of the previous 12 months and other customer features such as location, type of customer, voltage and whether there are meter reading notes during that period. On the test set, an accuracy of 0.8717, a precision of 0.6503 and a recall of 0.2947 are reported.
Consumption profiles of 5K Brazilian industrial customer profiles are analyzed in \cite{ramos2012identification}. Each customer profile contains 10 features including the demand billed, maximum demand, installed power, etc. In this setting, a SVM slightly outperforms $k$-nearest neighbors (KNN) and a neural network, for which test accuracies of 0.9628, 0.9620 and 0.9448, respectively, are reported. 
We have discussed the class imbalance and evaluation metric selection of NTL detection in \cite{glauner2016large} and shown that a large-scale machine learning approach outperforms rule-based Boolean and fuzzy logic expert systems. 
Covariate shift refers to the problem of training data (i.e. the set of inspection results) and production data (i.e. the set of customers to generate inspections for) having different distributions. We have shown in \cite{glauner2017big} that the sample of inspected customers may be biased, i.e. it does not represent the population of all customers. As a consequence, machine learning models trained on these inspection results may be biased as well and therefore may lead to unreliable predictions of whether customers cause NTL or not.
Furthermore, we have shown that the neighborhoods of customers yield significant information in order to decide whether a customer causes a NTL or not \cite{glauner2016neighborhood, meira2017distilling}.

In the literature, different approaches for visualization of NTL are reported. In order to support the decision making, the visualization of the network topology on feeder level as well as load curves on transformer level is proposed in \cite{abaide2010assessment}. In addition, the density of NTL in a 2D map is visualized in \cite{porras2015identification}. For analytics in power grids as a whole, the need for novel and more powerful visualization techniques is argued in \cite{trindade2016data}. The proposed approaches include heat maps and risk maps. All methods for visualization of NTL proposed in the literature focus only on 2D representations.

We are currently undergoing a paradigm shift in data visualization from not only 2D to 3D, but rather to augmented reality using holographic projections \cite{olshannikova2015visualizing}. This shift allows to better understand and experience data \cite{hoffman2016future}. Users are not constrained to looking at data on a screen, as they can interact with the data, e.g. walking around holograms to get a better understanding of Big Data sets. This comes with the benefit of increased productivity as users can use their hands to turn and manipulate objects rather than getting distracted caused by a change of focus from the screen to the input devices such as keyboards or mice \cite{lv2016managing}. A number of successful applications of holographic projections have been described in the literature including guided assembly instructions \cite{evans2017evaluating} as well as a  combination of different geographical information data sources in city management \cite{lv2016managing}. The literature also discusses the limitations of 3D visualizations, such as that users mistakenly may have greater confidence in the quality of the data \cite{zanola2009effect}.

\section{Detection of NTL}
\label{chapter:detection}
\subsection{Data}

The data used in this paper comes from an electricity provider in Brazil and consists of 3.6M customers. The data contains 820K inspection results, such as inspection date, presence of fraud or irregularity, type of NTL and inspection notes. 620K customers have been inspected at least once and the remaining \texttildelow 3M customers have never been inspected.
Third, there are 195M meter readings from 2011 to 2016 such as consumption in kWh, date of meter reading and number of days between meter readings.
From the 620K customers for which an inspection result is available, only the most recent inspection result is used in the experiments in Section~\ref{chapter:eval}. 
The available data per customer $m$ is a complete time series of monthly meter readings of electricity consumption in kWh over the last $N$ months before the most recent inspection, described as follows:
\begin{align*}
C^{(m)} = [C^{(m)}_0, ..., C^{(m)}_{N-1}],
\end{align*}
where $C^{(m)}_{N-1}$ is the most recent meter reading before the inspection.
For greater $N$, less customers with a complete time series are available. In contrast, for smaller $N$, less information per customer is available.

\begin{table*}[t]
\renewcommand{\arraystretch}{1.3}
\caption{Model Parameters.}
\label{table:params}
\centering
\begin{tabular}{|l|c|c|c|c|c|}
\hline
Parameter & Values & DT & RF & GBT & LSVM \\
\hline
\hline
Learning rate & $[0.0001, 1]$ (log space) & & & \checkmark & \\
\hline
Loss function & $\{$AdaBoost, deviance$\}$ & & & \checkmark & \\
\hline
Max. number of leaves & $[2, 1000)$ & \checkmark & \checkmark & \checkmark & \\
\hline
Max. number of levels & $[1, 50)$ & \checkmark & \checkmark & \checkmark & \\
\hline
Measure of the purity of a split & $\{$entropy, gini$\}$ & \checkmark & \checkmark & & \\
\hline
Min. number of samples required to be at a leaf & $[1, 1000)$ & \checkmark & \checkmark & \checkmark & \\
\hline
Min. number of samples required to split a node & $[2, 50)$ & \checkmark & \checkmark & \checkmark & \\
\hline
Number of estimators & $20$ & & \checkmark & \checkmark & \\
\hline
$L_2$ regularization & $$[0.001, 10] (log space) & & & &  \checkmark \\
\hline
\end{tabular}
\end{table*}

\subsection{Features}
In this section, we describe the features that we compute from a customer's consumption time series $C^{(m)}$ for the detection of NTL.

\subsubsection{Difference Features}
The intra year difference
\begin{align*}
\operatorname{intra\_year}_d^{(m)} = C_d^{(m)} - C_{d-K}^{(m)},
\end{align*}
for $K = 12$, is the change of consumption to the consumption in the same month of the previous year. In total, there are $N - 12$ intra year difference features.

The intra year seasonal difference
\begin{align*}
\operatorname{intra\_year\_seasonal}_d^{(m)} = C_d^{(m)} - \frac{1}{3} \times\sum^{d-K+1}_{k=d-K-1} C_{k}^{(m)},
\end{align*}
for $K = 12$, is the change of consumption to the mean of the same season in the previous year. In total, there are $N - 13$ intra year seasonal difference features.

The fixed interval
\begin{align*}
\operatorname{fixed\_interval}_d^{(m)} = C_d^{(m)} - \frac{1}{K} \times \sum^{d-1}_{k=d-K} C_{k}^{(m)},
\end{align*}
for $K \in \{3, 6, 12\}$, is the change of consumption to the mean consumption in a period of time directly before a meter reading. In total, there are $3\times (N - 12)$ fixed interval features.
These features are inspired by \cite{meira2017distilling}, in which they are proposed only for the most recent meter reading. Instead, we compute these features for the entire consumption time series.

\subsubsection{Daily Averages}
A daily average consumption feature during month $d$ for customer $m$ in kWh is:
\begin{align}
\operatorname{daily\_avg}_d^{(m)} = \frac{C_d^{(m)}}{R^{(m)}_{d} - R^{(m)}_{d-1}}.
\end{align}

$C_d^{(m)}$ is the consumption between the meter reading $R^{(m)}_d$ of month $d$ and the previous one $R^{(m)}_{d-1}$ in month $d-1$. $R^{(m)}_{d} - R^{(m)}_{d-1}$ is the number of days between both meter readings of customer $m$. In total, there are $N-1$ daily average consumption features.
This feature type is successfully used in a number of publications on NTL detection \cite{nagi2010nontechnical, nagi2011improving, glauner2016large, nagi2008non, nagi2008detection}. It is therefore also relevant to our research.

\subsubsection{Generic Time Series Features}
In order to catch more characteristics of the consumption time series, we compute 222 generic time series features from it, comprising:
\begin{itemize}
\item Summary statistics, such as maximum, variance or kurtosis.
\item Characteristics from sample distribution, such as absolute energy, whether a distribution is symmetric or the number of data points above the median.
\item Observed dynamics, such as fast Fourier transformation coefficients, autocorrelation lags or mean value of the second derivative.
\end{itemize}

The full list of features is provided in \cite{christ2016distributed}. 

\subsection{Feature Selection}
In total, 304 features are computed. In the subsequent learning phase, only the meaningful features should be used. One common dimensionality reduction method is the principal component analysis (PCA). However, time series, and in particular real-world data sets, are noisy, which can lead to poor performance of PCA \cite{fu2011review}. It is for that reason that we do not use PCA for the feature selection. Instead, we employ hypothesis tests to the features in order to retain the ones that are statistically relevant. These tests are based on the assumption that a feature $x_k$ is meaningful for the prediction of the binary label vector $y$ if $x_k$ and $y$ are not statistically independent \cite{radivojac2004feature}. For binary features, we use Fisher's exact test \cite{fisher1922interpretation}. In contrast, for continuous features, we use the Kolmogorov-Smirnov test \cite{massey1951kolmogorov}.

\subsection{Classifiers}
\subsubsection{Decision Tree}
Decision tree learners such as ID3 or C4.5 \cite{quinlan1993c4} recursively split the input space by choosing the remaining most discriminative feature of a data set. To predict, the learned tree is traversed top-down.

\subsubsection{Random Forest}
A random forest \cite{ho1995random} is an ensemble estimator that comprises a number of decision trees. Each tree is trained on a subsample of the data and feature set in order to control overfitting. In the prediction phase, a majority vote is made of the predictions of the individual trees.

\subsubsection{Gradient Boosted Tree}
A gradient boosted tree \cite{chen2016xgboost} is also an ensemble of decision trees. The ensemble is boosted by combining weak classifiers (i.e. classifiers that work little better than a random guess) into a strong one. The ensemble is built by optimizing a loss function.

\subsubsection{Support Vector Machine}
A support vector machine (SVM) \cite{vapnik1999overview} is a maximum margin classifier, i.e. it creates a maximum separation between classes.
Support vectors hold up the separating hyperplane. In practice, they are just a small fraction of the training examples.
Therefore, a SVM is often less prone to overfitting than other classifiers, such as a neural network \cite{cao2003support}.
The training of a SVM can be defined as a Lagrangian dual problem having a convex cost function.
By default, the separating hyperplane is linear.
Training of SVMs using a kernel to map the input to higher dimension is only feasible for several dozens of thousands of training examples in a realistic amount of time \cite{CC01a}.
Therefore, for Big Data sets only a linear implementation of SVMs is practically usable \cite{scikit-learn}.

\begin{table}[!t]
\renewcommand{\arraystretch}{1.3}
\caption{Number of Features before and after Selection.}
\label{table:features}
\centering
\begin{tabular}{|l|c|c|}
\hline
Name & \#Features & \#Retained features \\
\hline
\hline
Daily average (AVG) & 23 & 18 \\
\hline
Fixed interval & 36 & 34 \\
\hline
Generic time series (GTS) & 222 & 162 \\
\hline
Intra year difference & 12 & 12 \\
\hline
Intra year seasonal difference & 11 & 11 \\
\hline
\hline
Total & 304 & 237 \\
\hline
\end{tabular}
\end{table}

\begin{table*}[!ht]
\renewcommand{\arraystretch}{1.3}
\setlength{\tabcolsep}{0pt}
\caption{Test Performance of Classifiers on Features from Measured Consumption Data.}
\begin{minipage} {0.95\textwidth}
\label{table:resCons}
\begin{center}
\begin{tabular}{|l|c|c|c|c|c|c|c|c|c|c|c|c|c|c|}
\hline
\multirow{ 2}{*}{Clf.} & \multicolumn{2}{|c|}{GTS} & \multicolumn{2}{|c|}{AVG} & \multicolumn{2}{|c|}{DIF} & \multicolumn{2}{|c|}{GTS+AVG} & \multicolumn{2}{|c|}{GTS+DIF} & \multicolumn{2}{|c|}{AVG+DIF} & \multicolumn{2}{|c|}{GTS+AVG+DIF} \\
\cline{2-15}
 & $X_{all}$ & $X_{ret}$ & $X_{all}$ & $X_{ret}$ & $X_{all}$ & $X_{ret}$ & $X_{all}$ & $X_{ret}$ & $X_{all}$ & $X_{ret}$ & $X_{all}$ & $X_{ret}$ & $X_{all}$ & $X_{ret}$ \\
\hline
\hline
DT & 0.64544 & \cellcolor[gray]{0.9} 0.64625 & \cellcolor[gray]{0.9} 0.64037 & 0.63985 & 0.63730 & \cellcolor[gray]{0.9} 0.63792 & \cellcolor[gray]{0.9} 0.64712 & 0.64705 & 0.64638 & \cellcolor[gray]{0.9} 0.64647 & \cellcolor[gray]{0.9} 0.64348 & 0.64312 & 0.64646 & \cellcolor[gray]{0.9} $0.64765^{f}$ \\
\hline
RF & $0.65665^{c}$ & \cellcolor[gray]{0.9} $0.65726^{c}$ & $0.65083^{c}$ & \cellcolor[gray]{0.9} $0.65248^{c}$ & \cellcolor[gray]{0.9} $0.65529^{c}$ & $0.65459^{c}$ & $0.65800^{c}$ & \cellcolor[gray]{0.9} $0.65835^{c}$ & \cellcolor[gray]{0.9} $0.65911^{c}$ & $0.65896^{c}$ & \cellcolor[gray]{0.9} $0.65858^{c}$ & $0.65755^{c}$ & $0.65747^{c}$ & \cellcolor[gray]{0.9} $\textbf{0.65977}^{cf}$ \\
\hline
GBT & \cellcolor[gray]{0.9} 0.63149 & 0.63125 & \cellcolor[gray]{0.9}  0.63234 & 0.63186 & 0.62869 & \cellcolor[gray]{0.9}  0.63019 & 0.63262 & \cellcolor[gray]{0.9} 0.63322 & 0.63319 & \cellcolor[gray]{0.9} $0.63358^{f}$ & \cellcolor[gray]{0.9} 0.63261 & 0.63245 & 0.63354 & \cellcolor[gray]{0.9} 0.63355 \\
\hline
LSVM & \cellcolor[gray]{0.9} 0.63696 & 0.63656 & \cellcolor[gray]{0.9} 0.54982 & 0.54933 & 0.55749 & \cellcolor[gray]{0.9} 0.55843 & 0.63725 & \cellcolor[gray]{0.9} 0.63689 & \cellcolor[gray]{0.9} 0.63731 & 0.63693 & 0.57173 & \cellcolor[gray]{0.9} 0.57432 & 0.63728 & \cellcolor[gray]{0.9} $0.63760^{f}$ \\
\hline
\end{tabular}
\end{center}
Test AUC for combinations of decision tree (DT), random forest (RF), gradient boosted tree (GBT) and linear support vector machine (LSVM) classifiers trained on sets composed of general time series (GTS), daily average (AVG) and difference (DIF) features.\\
The best overall combination of classifier and feature set is \textbf{highlighted}.\\
Per combination of classifier and feature set, the better result on either a full feature set ($X_{all})$ or retained feature set ($X_{ret})$ is \colorbox{light-gray}{highlighted}. \\
$^c$ denotes the best classifier per feature set.\\
$^f$ denotes the best feature set per classifier.
\end{minipage}
\end{table*}

\section{Evaluation}
\label{chapter:eval}

\subsection{Metric}
The performance measure used in the following experiments is the area under the receiver-operating curve (AUC) \cite{van2003area}. It plots the true positive rate or recall against the false positive rate. It is particularly useful for NTL detection, as it allows to handle imbalanced datasets and puts correct and incorrect inspection results in relation to each other.
The superiority of the AUC over other metrics such as accuracy, precision or recall with respect to the problem of NTL detection has been witnessed in the literature \cite{glauner2016large}.

\subsection{Experimental Setup}
We experimentally determined $N=24$ months to work the best for the following experiments. Using $N=24$ allows the consumption data to reflect seasonality in the experiments. As a consequence, $M=150,700$ customers are retained for the experiments.
This data set is imbalanced: 100,471 have a negative label (non-NTL), whereas 50,229 have a positive one (NTL). Therefore, 33.33\% of the customers used in the following experiments have been found to cause NTL.

We train the decision tree (DT), random forest (RF), gradient boosted tree (GBT) and linear support vector machine (LSVM) classifiers as follows:
\begin{itemize}
\item Handling class imbalance: We handle the class imbalance during training by assigning class weights to the examples of both classes in the training set:
\begin{align}
w_0 &= \frac{\#\operatorname{examples}}{\#\operatorname{examples}_{C=0}},\\
w_1 &= \frac{\#\operatorname{examples}}{\#\operatorname{examples}_{C=1}}.
\end{align}
\item Performing model selection: We want to find the model which is able to distinguish between NTL and non-NTL customers the best.
For this, we optimize various parameters for every classifier. The complete list of parameters and considered values per classifier is depicted in Table~\ref{table:params}. We use randomized grid search, which samples from the joint distribution of model parameters. In contrast to grid search, randomized grid search does not try out all parameter values. We use 100 sampled models in every model selection.
\item Handling overfitting: We also employ model selection that splits the data set into $k=10$ folds. This leads to a more reliable model for NTL detection. The AUC reported per model is the average of the AUCs of the $k$ test sets.
\end{itemize}

\subsection{Implementation}
All computations were run on a server with 80 cores and 128 GB of RAM. 
The entire code was implemented in Python using \texttt{scikit-learn} \cite{scikit-learn} for machine learning. \texttt{scikit-learn} allows to distribute the training of the numerous classifiers among all cores.  Using this infrastructure, the extraction of features took 6 hours. The feature selection took only 1 minute. The extensive model selection of classifiers took 4 days. In deployment, the training of classifiers will perform significantly faster as the extensive model selection needs to be performed only when a new data set is used. We have also noticed that about 90\% of the training time was spent on the gradient boosted tree. Therefore, a significant speedup can be achieved in deployment when skipping the training of this classifier.

\subsection{Feature Selection}
We first compute the features described in Section~\ref{chapter:detection} and then perform the feature selection. In summary, there are three types of features: (1) generic time series (GTS) features, (2) daily average features (AVG) and (3) difference features (DIF) composed of fixed interval, intra year difference and intra year seasonal difference features. 
The numbers of features before and after selection are depicted in Table~\ref{table:features}. In total, 237 out of the 304 features are retained. The relevance of our hand-crafted difference features is confirmed: All intra year difference and intra year seasonal difference features are retained. In addition, 34 out of 36 fixed interval features are retained. The 2 features are not retained for $K=3$, which is most likely due to the too short span of time they reflect.
As a matter of fact, daily average features are widely used in the research literature on NTL detection. However, only 18 out of 23 daily average consumption features (i.e. 78\%) are retained. The 5 daily average consumption features that are not retained are the ones for the first - i.e. the oldest - 6 months of the 24-month window. The statistical feature check leads to the conclusion that this type of feature is only useful for about 1.5 years of our data for NTL detection.
In addition, 73\% of the generic time series features are retained after the statistical relevance check. As these features are generic and not particularly made for NTL detection, it is to no surprise that the retention rate for these features is the lowest.

\subsection{Classification Results}
We train the four classifiers on each of the GTS, AVG and DIF feature sets as well as on all combinations thereof. The test performance of the best model per experiment returned by the model selection is depicted in Table~\ref{table:resCons}. The best test AUC of 0.65977 is achieved for training the random forest classifier on the combination of the retained GTS, AVG and DIF features. In general, the random forest classifier works the best for every feature set.
In total, we report the results of 28 experiments in Table~\ref{table:resCons}, both for the full feature sets as well as the retained feature sets. In 16 experiments, the feature selection leads to better results over using all features. Our observation can be explained by the ``no free lunch theorem", which states that no model is generally better than others \cite{wolpert1996lack}. However, our best result of 0.65977 is achieved for the retained feature set.

Generally, we observe that a combination of two or three feature sets leads to a better test result than for any of the respective single feature sets. An example to demonstrate this observation is as follows: The random forest classifier achieves test AUCs of 0.65726, 0.65248 and 0.65459 for the retained GTS, AVG and DIF features, respectively. It then achieves test AUCs of 0.65835, 0.65896, 0.65755 and 0.65977 for the retained GTS+AVG, GTS+DIF, AVG+DIF and GTS+AVG+DIF feature sets, respectively. Therefore, the test AUCs for each of the combined feature sets are greater than the test AUCs for any of the single feature sets. 

\subsection{Discussion}
Previous works that employ the widely-used daily average features established a baseline that only achieved an AUC of slightly above 0.5 \cite{nagi2010ntl,glauner2016large}, i.e. slightly above chance, on real-world NTL detection data sets using linear classifiers.
First and foremost, we want to highlight that increasing the performance of machine learning models on noisy real-world data sets is far more challenging than doing so on academic data sets that were created and curated in controlled environments. Furthermore, a small increase of the performance of a real-world model can lead to a major increase of the market value of a company. Our framework presented in this paper significantly outperforms the baselines established in the literature. As a consequence, our models lead to a better detection of NTL and thus to an increase of revenue and profit for electricity providers as well as an increase of stability and reliability in their critical infrastructure.
Our NTL detection framework allows other electricity providers to apply our extensive feature extraction, feature selection and model selection techniques to their data sets, which can lead to potentially greater improvements of NTL detection in their power networks.

It is to our surprise that the gradient boosted tree classifier performs consistently worse than the random forest classifier in our experiments. In the literature, the gradient boosted tree is reported to often lead in a wide range of classification problems \cite{chen2016xgboost}.
However, our observation can also be explained by the ``no free lunch theorem".

\section{Holographic Visualization of NTL}
\label{chapter:visualization}
The NTL detection approach presented in Section~\ref{chapter:detection} and evaluated in Section~\ref{chapter:eval} allows to predict whether customers cause NTL or not. It can then be used to trigger possible inspections of customers that have irregular electricity consumption patterns. Subsequently, technicians carry out inspections, which allow them to remove possible manipulations or malfunctions of the power distribution infrastructure. Furthermore, the fraudulent customers can be charged for the additional electricity consumed. Generally, carrying out inspections is costly, as it requires physical presence of technicians.
In order to increase both the ROI of the limited number of inspections and the reliability and stability of the power grid, electricity providers in practice strongly rely on expert knowledge for making the decision of whether to inspect a customer or not \cite{glauner2016large}. As a consequence, electricity providers are reluctant to move to large-deployments of NTL detection systems based on machine learning.
We therefore aim to combine automated statistical decision making for generating inspection proposals with incorporating knowledge of the domain experts at the electricity providers for making the final decisions of which customers to inspect.

\begin{figure}[!t]
\centering
\includegraphics[width=0.5\textwidth]{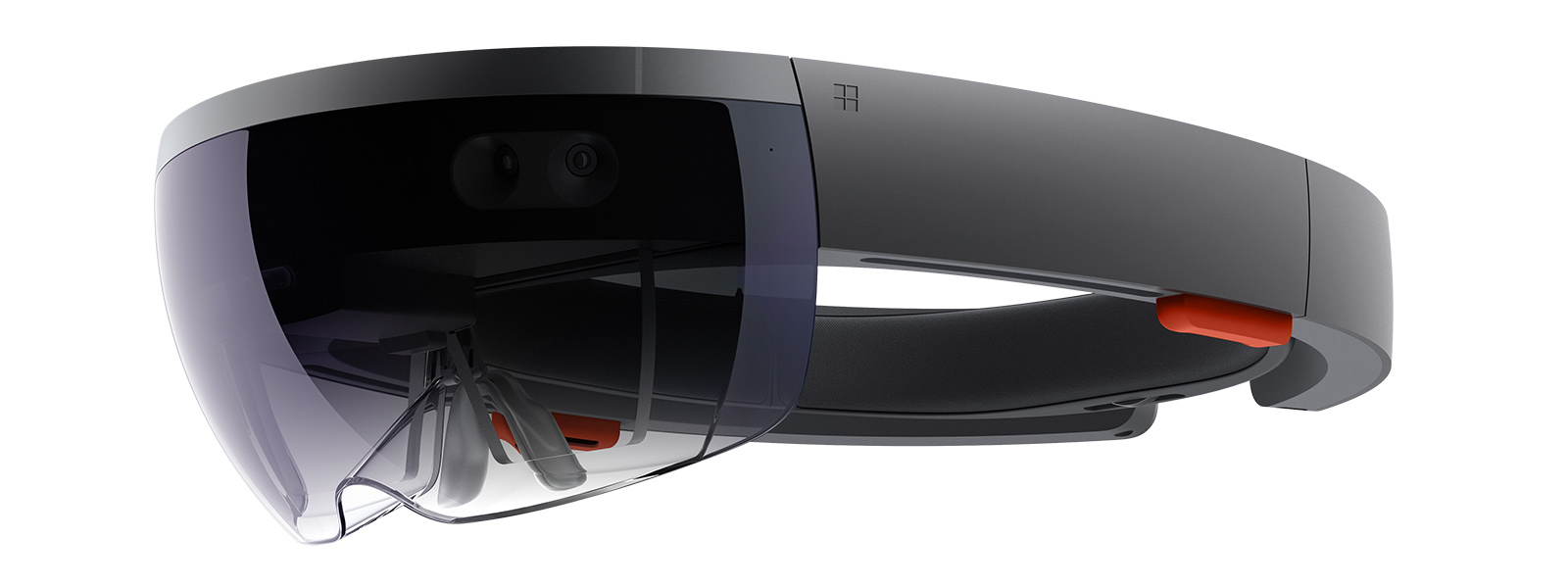}
\caption[XXX]{Microsoft HoloLens \cite{hololens2016}.}
\label{fig:HoloLens}
\end{figure}

\subsection{HoloLens}
Mixed reality smartglasses such as the Microsoft HoloLens \cite{hololens2016} depicted in Figure~\ref{fig:HoloLens} allow users to combine holographic projections with the real world.
The Hololens offers their user a new perception of 3D models and, perhaps, can provide a new meaning to it. Visualization of data through holograms has found its application in many areas. In medicine, future doctors can study human anatomy by looking at a representation of the human body and navigate through muscles, organs and skeletons \cite{taylor2016creating}.
The HoloLens has the ability to perform the holoportation, which allows to virtually place users to remote locations to see, hear and interact with others.
Users can walk around holograms and interact with them using gaze, gestures or voice in the most natural way. Spatial sound allows hearing holograms even if they are behind the user, considering its position and direction of the sound. Spatial mapping features provide a real-world representation of surfaces, creating convincing holograms in augmented reality.

\begin{figure}[!t]
\centering
\includegraphics[width=0.5\textwidth]{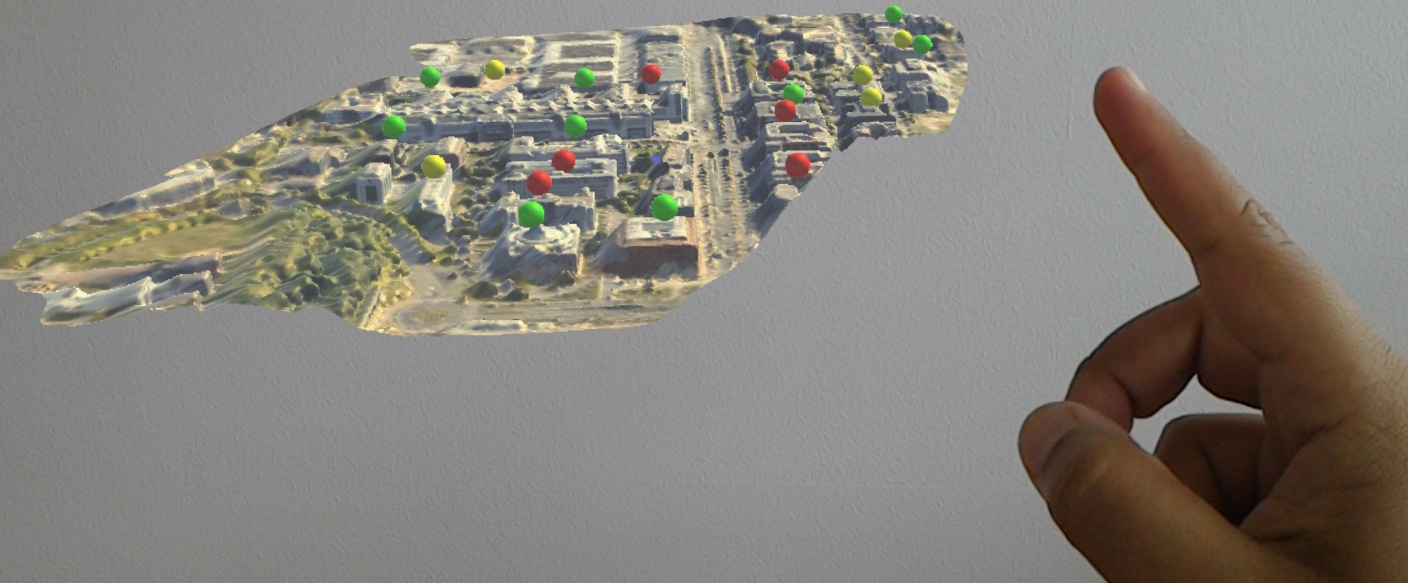}
\caption[XXX]{Gesture interactions with the spatial hologram allow to select customers as well as to zoom into or rotate holograms. We also provide a future yellow label that depicts a borderline case, which requires a manual check by domain experts.}
\label{fig:holo1}
\end{figure}

\begin{figure}[!t]
\centering
\includegraphics[width=0.5\textwidth]{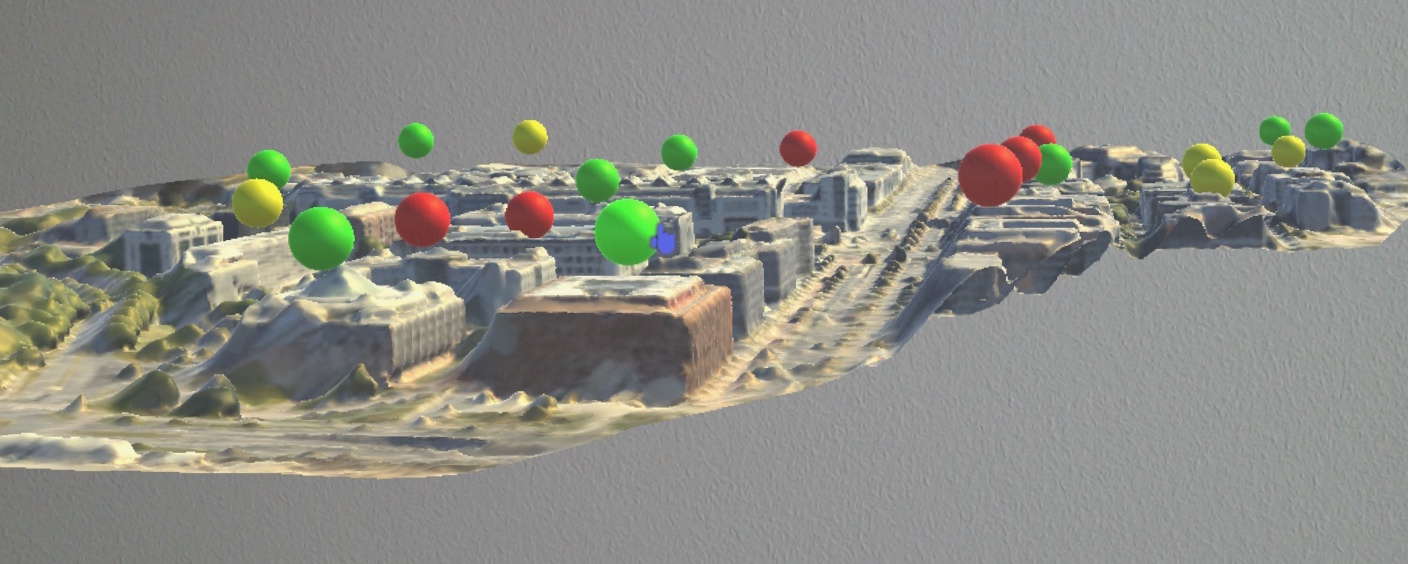}
\caption[XXX]{Zoomed and rotated view on the spatial hologram.}
\label{fig:holo2}
\end{figure}

\subsection{Implementation}
We created a 3D model using Google Earth Pro and Blender. Our model allows us to visualize customers and their neighborhood in a 3D spatial hologram that is depicted in Figure~\ref{fig:holo1}. 
A movie was recorded to capture the scene and all its objects from the different angles through Google Earth Pro. Afterwards, images were extracted in Windows Movie Maker from that movie at the best experimentally determined rate of 1 frame/sec. Then, those images were loaded in Blender, which in turn created a 3D FBX model. This model was exported to Unity. Holographic effects were implemented through \texttt{HoloToolkit-Unity} \cite{holotoolkit2017}. The \texttt{GameObject}s that handle input events implement the \texttt{IInputHandler} interface for tap and hold gestures. Classes that implement the \texttt{IManipulationHandler} interface handle manipulation gestures such as moving and rotating actions.

\subsection{Results}
This application is used by domain experts at the electricity providers who perceive that customers are classified as either regular (green) or irregular (red). Domain experts can walk around a spatial hologram and observe the data from different directions. Using their hand, they can also interact with the hologram in different ways, such as zooming into or rotating the hologram as depicted in Figure~\ref{fig:holo2}.

\begin{figure}[!t]
\centering
\includegraphics[width=0.5\textwidth]{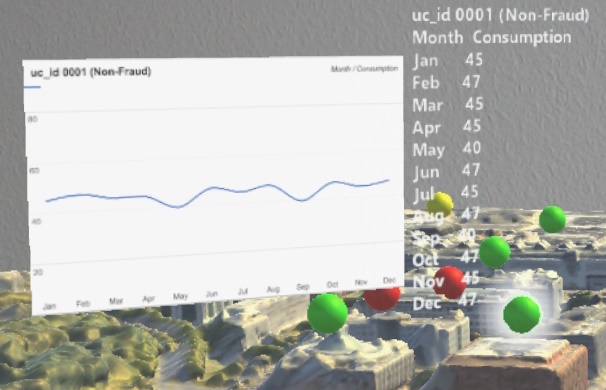}
\caption[XXX]{Detailed view of a customer depicted by a green dot predicted to have a regular power consumption pattern.}
\label{fig:holo3}
\end{figure}

\begin{figure}[!t]
\centering
\includegraphics[width=0.5\textwidth]{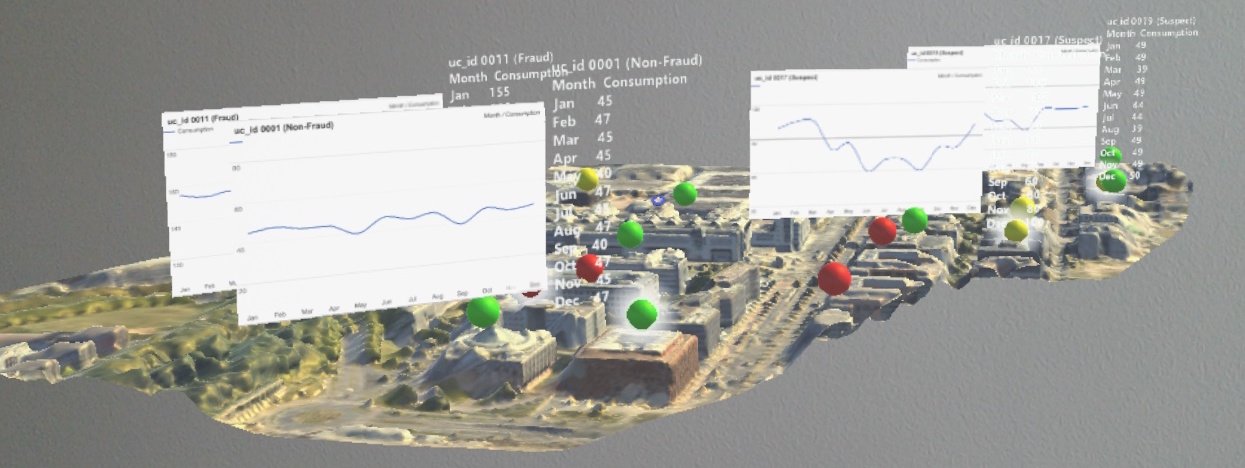}
\caption[XXX]{Multi-view on multiple customers' power consumption history.}
\label{fig:holo4}
\end{figure}

Domain experts can also learn more about a customer by tapping on it with their finger. The spatial hologram then also depicts the consumption profile of the respective customer over a selected period of time such as the previous 12 months. A customer with a predicted regular consumption profile is depicted in Figure~\ref{fig:holo3}. This customer's consumption has only changed very little in the last 12 months. As a consequence, the machine learning system classified this customer as non-NTL (green). A customer with an irregular consumption profile is depicted in Figure~\ref{fig:example}. This customer's consumption has undergone a significant drop over the last few months. Therefore, the machine learning system classified this customer as NTL (red).
In both cases, domain experts can compare their observations with the prediction made by the machine learning system. If the prediction is not plausible, domain experts can choose not to follow the recommendation and therefore decide whether to inspect a customer.
Our visualization allows domain experts to take the neighborhood of customers into account in order to decide which customers to inspect. Aside from the actual spatial visualization of satellite images of a neighborhood, domain experts can also visualize the consumption profile of neighbors as visualized in Figure~\ref{fig:holo4} for comparing customers in order to decide whether to inspect a customer.

\subsection{Discussion}
Our holographic spatial visualization of customers and their neighborhood comes with the benefit of increased productivity. It has previously been shown that the neighborhoods of customers yield significant information in order to decide whether a customer causes a NTL or not \cite{glauner2016neighborhood, meira2017distilling}. There are many  interpretations of this fact. For example, fraudulent customers may share their knowledge with neighbors or there may be a correlation between electricity theft and the level of prosperity of a neighborhood.
Our system allows to increase the ROI of inspections as well as to increase both the reliability and stability of the power grid by incorporating expert knowledge in the decision making process. Also, domain experts can use their hands to turn and manipulate objects rather than getting distracted by a change of focus from the screen to the input devices such as a keyboard or mouse.

\section{Conclusion and Future Work}
\label{chapter:end}
In this work, we have proposed a novel system for detecting non-technical losses (NTL) for a real-world data set of 3.6M customers. In the first stage, a machine learning system learns to predict whether a customer causes NTL or not. In order to do so, we have proposed to extract a number of domain-specific features from the noisy consumption data. We have shown the statistical relevance of these features over generic time series features. As a consequence, our machine learning system allows to detect NTL better than previous works described in the literature.
In the second stage, we put the prediction results into context by visualizing further data of the customers and their neighborhoods in a spatial hologram using a Microsoft HoloLens. Using this hologram, domain experts can then review and amend the suggestions of which customers to inspect. As a result, they can make the final decisions of which customers to inspect in order to increase the ROI of the limited number of inspections.

We have previously referred to the main challenges to solve in order to advance NTL detection. We believe that covariate shift is one of the main impediments in advancing NTL detection. It has been argued that covariate shift is currently one of the main impediments in a wide range of real-world Big Data machine learning problems \cite{l2017machine}. Therefore, reducing the covariate shift in the data should be a future priority in the detection of NTL. We therefore expect our models to perform better after the reduction of covariate shift.
In order to make the visualization ready for large-scale production, we are planning to integrate other 3D data sources as well as to add compression algorithms such that large maps can be transferred to the HoloLens. We are also interested in visualizing other quantities, such as prosperity levels or credit worthiness, in the spatial holograms.
We are also interested in exploring unsupervised methods in the future in order to build system that perform with only the availability of consumption data.

\section*{Acknowledgment}
We would like to thank Tamer Aidek and Yves Rangoni from CHOICE Technologies Holding S\`arl for contributing many good ideas to our discussions. This work has been partially funded by the Luxembourg National Research Fund.



\bibliographystyle{IEEEtran}
\bibliography{references}
%

\end{document}